\documentclass[runningheads]{llncs}
\usepackage{graphicx}
\usepackage{amsmath,amssymb} 
\usepackage{color}
\usepackage[width=122mm,left=12mm,paperwidth=146mm,height=193mm,top=12mm,paperheight=217mm]{geometry}
\begin{document}
\pagestyle{headings}
\mainmatter

\title{Deep Kinematic Pose Regression} 

\titlerunning{Deep Kinematic Pose Regression}

\authorrunning{Zhou et al.}

\author{Xingyi Zhou$^1$, Xiao Sun$^2$, Wei Zhang$^1$,  Shuang Liang$^3$\thanks{Corresponding author.}, Yichen Wei$^2$}


\institute{$^1$Shanghai Key Laboratory of Intelligent Information Processing \\
School of Computer Science, Fudan University \\
	$^2$ Microsoft Research\\
	$^3$ Tongji University\\
	\email{ \{zhouxy13,weizh\}@fudan.edu.cn, \{xias, yichenw\}@microsoft.com, shuangliang@tongji.edu.cn}
}

\maketitle

\begin{abstract}
Learning articulated object pose is inherently difficult because the pose is high dimensional but has many structural constraints. Most existing work do not model such constraints and does not guarantee the geometric validity of their pose estimation, therefore requiring a post-processing to recover the correct geometry if desired, which is cumbersome and sub-optimal. In this work, we propose to directly embed a kinematic object model into the deep neutral network learning for general articulated object pose estimation. The kinematic function is defined on the appropriately parameterized object motion variables. It is differentiable and can be used in the gradient descent based optimization in network training. The prior knowledge on the object geometric model is fully exploited and the structure is guaranteed to be valid. We show convincing experiment results on a toy example and the 3D human pose estimation problem. For the latter we achieve state-of-the-art result on Human3.6M dataset.
\end{abstract}

\section{Introduction}

Estimating the pose of objects is important for understanding the behavior of the object and relevant high level tasks, e.g., facial point localization for expression recognition, human pose estimation for action recognition. It is a fundamental problem in computer vision and has been heavily studied for decades. Yet, it remains challenging, especially when object pose and appearance is complex, e.g., human pose estimation from single view RGB images.

There is a vast range of definitions for object pose. In the simple case, the pose just refers to the global viewpoint of rigid objects, such as car~\cite{Zhu_2015_ICCV} or head~\cite{Meyer_2015_ICCV}. But more often, the pose refers to a set of semantically important points on the object (rigid or non-rigid). The points could be landmarks that can be easily distinguished from their appearances, e.g., eyes or nose on human face~\cite{Jourabloo_2016_CVPR}, and wings or tail on bird~\cite{yu2016deep}. The points could further be the physical joints that defines the geometry of complex articulated objects, such as human hand~\cite{zhou2016model,oberweger2015hands} and human body~\cite{li20143d,Zhou_2016_CVPR,Tekin_2016_CVPR}.

\begin{figure}
\begin{center}
\includegraphics[width=0.95\linewidth]{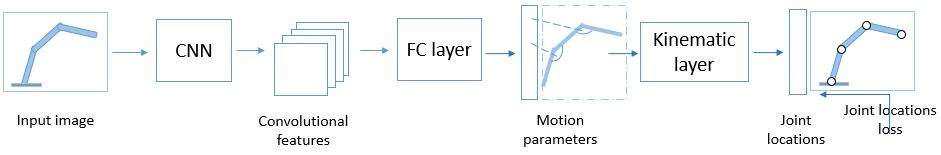}
\end{center}
   \caption{Illustration of our framework. The input image undergoes a convolutional neutral network and a fully connected layer to output model motion parameters (global potision and rotation angles). The kinematic layer maps the motion parameters to joints. The joints are connected to ground truth joints to compute the joint loss that drives the network training.}
\label{fig:framework}
\end{figure}

Arguably, the articulated object pose estimation is the most challenging. Such object pose is usually very high dimensional and inherently structured. How to effectively represent the pose and perform structure-preserving learning is hard and have been heavily studied. Some approaches represent the object pose in a non-parametric way (as a number of points) and directly learn the pose from data~\cite{sun2015cascaded,shotton2013efficient,Carreira_2016_CVPR}. The inherent structure is implicitly learnt and modeled from data. Many other approaches use a low dimensional representation by using dimensionality reduction techniques such as PCA~\cite{ionescu2014human3,oberweger2015hands}, sparse coding~\cite{Wang_2014_CVPR,Zhou_2015_CVPR,Zhou_2016_CVPR} or auto-encoder~\cite{tekin2016structured}. The structure information is embedded in the low dimensional space. Yet, such embedding is mostly linear and cannot well preserve the complex articulated structural constraints.

In this work, we propose to directly incorporate the articulated object model into the deep neutral network learning, which is the dominant approach for object pose estimation nowadays, for hand~\cite{tompson14tog,supancic2015depth,oberweger2015hands,oberweger2015training,zhou2016model,Ge_2016_CVPR} or human body\cite{toshev2014deeppose,Wei_2016_CVPR,DBLP:journals/corr/NewellYD16,insafutdinov2016deepercut,li20143d,Li_2015_ICCV,Tekin_2016_CVPR,tekin2016structured,Zhou_2016_CVPR}. Our motivation is simple and intuitive. The kinematic model of such objects is well known as prior knowledge, such as the object bone lengths, bone connections and definition of joint rotations. From such knowledge, it is feasible to define a continuous and differentiable kinematic function with respect to the model motion parameters, which are the rotation angles. The kinematic function can be readily put into a neutral network as a special layer. The standard gradient descent based optimization can be performed in the same way for network training. The learning framework is exemplified in Fig.~\ref{fig:framework}. In this way, the learning fully respects the model geometry and preserves the structural constraints. Such end-to-end learning is better than the previous approaches that rely on a separate post-processing step to recover the object geometry~\cite{tompson14tog,Zhou_2016_CVPR}.

This idea is firstly proposed in the recent work~\cite{zhou2016model} for depth based hand pose estimation and is shown working well. However, estimating 3D structure from depth is a simple problem by nature. It is still unclear how well the idea can be generalized to other objects and RGB images. In this work, we apply the idea to more problems (a toy example and human pose estimation) and for the first time show that the idea works successfully on different articulated pose estimation problems and inputs, indicating that the idea works in general. Especially, for the challenging 3D human pose estimation from single view RGB images, we present state-of-the-art results on the Human3.6M dataset~\cite{h36m_pami}.

\section{Related Work}
The literature on pose estimation is comprehensive. We review previous work from two perspectives that are mostly related to our work: object pose representation and deep learning based human pose estimation.

\subsection{Pose Representation}
An object pose consists of a number of related points. The key for pose representation is how to represent the mutual relationship or structural constraints between these points. There are a few different previous approaches.

\textbf{Pictorial Structure Model}
Pictorial structure model~\cite{felzenszwalb2005pictorial} is one of the most popular methods in early age. It represents joints as vertexes and joint relations as edges in a non-circular graph. Pose estimation is formulated as inference problems on the graph and solved with certain optimization algorithms. Its extensions~\cite{johnson2011learning,yang2011articulated,Pishchulin_2013_CVPR} achieve promising results in 2D human estimation, and has been extended to 3D human pose~\cite{Belagiannis_2014_CVPR}. The main drawback is that the inference algorithm on the graph is usually complex and slow.

\textbf{Linear Dictionary}
A widely-used method is to denote the structural points as a linear combination of templates or basis~\cite{Wang_2014_CVPR,Zhou_2015_CVPR,Zhou_2016_CVPR,Jourabloo_2016_CVPR}. ~\cite{Jourabloo_2016_CVPR} represent 3D face landmarks by a linear combination of shape bases~\cite{paysan20093d} and expression bases~\cite{cao2014facewarehouse}. It learns the shape, expression coefficients and camera view parameters alternatively. ~\cite{Wang_2014_CVPR} express 3D human pose by an over-complex dictionary with a sparse prior, and solve the sparse coding problem with alternating direction method. ~\cite{Zhou_2015_CVPR} assign individual camera view parameters for each pose template. The sparse representation is then relaxed to be a convex problem that can be solved efficiently.

\textbf{Linear Feature Embedding}
Some approaches learn a low dimensional embedding~\cite{ionescu2014human3,oberweger2015hands,h36m_pami,tekin2016structured} from the high dimensional pose. ~\cite{ionescu2014human3} applies PCA to the labeled 3D points of human pose. The pose estimation is then performed in the new orthogonal space. The similar idea is applied to 3D hand pose estimation~\cite{oberweger2015hands}. It uses PCA to project the 3D hand joints to a lower space as a physical constraint prior for hand. ~\cite{tekin2016structured} extend the linear PCA projector to a multi-layer anto-encoder. The decoder part is fine-tuned jointly with a convolutional neural network in an end-to-end manner. A common drawback in above linear representations is that the complex object pose is usually on a non-linear manifold in the high dimensional space that cannot be easily captured by a linear representation.

\textbf{Implicit Representation by Retrieval}
Many approaches~\cite{Choi_2015_ICCV,Li_2015_ICCV,Yasin_2016_CVPR} store massive examples in a database and perform pose estimation as retrieval, therefore avoiding the difficult pose representation problem. ~\cite{Choi_2015_ICCV} uses a nearest neighbors search of local shape descriptors. ~\cite{Li_2015_ICCV} proposes a max-margin structured learning framework to jointly embed the image and pose into the same space, and then estimates the pose of a new image by nearest neighbor search in this space. ~\cite{Yasin_2016_CVPR} builds an image database with 3D and 2D annotations, and uses a KD-tree to retrieve 3D pose whose 2D projection is similar to the input image. The performance of these approaches highly depends on the quality of the database. The efficiency of nearest neighbor search could be an issue when the database is large.

\textbf{Explicit Geometric Model} The most aggressive and thorough representation is to use an explicit and generative geometric model, including the motion and shape parameters of the object~\cite{toby15,bogo2016smpl}. Estimating the parameters of the model from the input image(s) is performed by heavy optimization algorithms. Such methods are rarely used in a learning based manner. The work in~\cite{zhou2016model} firstly uses a generative kinematic model for hand pose estimation in the deep learning framework. Inspire by this work, we extend the idea to more object pose estimation problems and different inputs, showing its general applicability, especially for the challenging problem of 3D human pose estimation from single view RGB images.

\subsection{Deep Learning on Human Pose Estimation}
The human pose estimation problem has been significantly advanced using deep learning since the pioneer deep pose work~\cite{toshev2014deeppose}. All current leading methods are based on deep neutral networks. ~\cite{Wei_2016_CVPR} shows that using 2D heat maps as intermediate supervision can dramatically improve the 2D human part detection results. ~\cite{DBLP:journals/corr/NewellYD16} use an hourglass shaped network to capture both bottom-up and top-down cues for accurate pose detection. ~\cite{insafutdinov2016deepercut} shows that directly using a deep residual network (152 layers)~\cite{He_2016_CVPR} is sufficient for high performance part detection. To adopt these fully-convolutional based heat map regression method for 3D pose estimation, an additional model fitting step is used~\cite{Zhou_2016_CVPR} as a post processing. Other approaches directly regress the 2D human pose~\cite{toshev2014deeppose,Carreira_2016_CVPR} or 3D human pose ~\cite{li20143d,tekin2016structured,Tekin_2016_CVPR}. These detection or regression based approaches ignore the prior knowledge of the human model and does not guarantee to preserve the object structure. They sometimes output geometrically invalid poses.

To our best knowledge, for the first time we show that integrating a kinematic object model into deep learning achieves state-of-the-art results in 3D human pose estimation from single view RGB images.

\section{Deep Kinematic Pose Estimation}
\subsection{Kinematic Model}

\begin{figure}
\begin{center}
\includegraphics[width=0.95\linewidth]{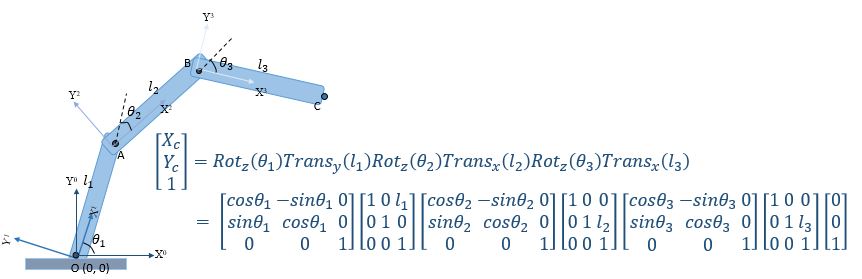}
\end{center}
   \caption{A sample 2D kinematic model. It has 3 and 4 joints. The joint location is calculated by multiplying a series of transformation matrices.}
\label{fig:kinematic}
\end{figure}

An articulated object is modeled as a kinematic model. A kinematic model is composed of several \emph{bones} and \emph{joints}. A bone is a segment of a fixed length, and a joint is the end point of a bone. One bone meets at another at a joint, forming a tree structure. Bones can rotate among a conjunct joint. Without loss generality, one joint is considered as the root joint (For example, wrist for human hand and pelvis for human body). The root defines the global position and global orientation of the object.

For a kinematic model of $J$ joints, it has $J - 1$ bones. Let $\{l_i\}_{i = 1} ^ {J - 1}$ be the collection of bone lengths, they are fixed for a specific subject and provided as prior knowledge. For different subjects, we assume they only differ in a global scale, i.e. $ \forall i, l_i' = s \times l_i$. The scale is also provided as prior knowledge, e.g. through a calibration process.

Let the rotation angle of the $i$-th joint be $\theta_i$, the motion parameter $\Theta$ includes the global position $\textbf{p}$, global orientation $\textbf{o}$, and all the rotation angles, $\Theta = \{\textbf{p}, \textbf{o}\} \cup \{\theta_i\}_{i = 1} ^ J$. The forward kinematic function is a mapping from motion parameter space to joint location space.
\begin{equation} \label{eq:map}
\mathcal{F}: \{\Theta\} \rightarrow \mathcal{Y}
\end{equation}
where $\mathcal{Y}$ is the coordinate for all joints, $\mathcal{Y} \in \mathcal{R} ^ {3 \times J}$ for 3D object and $\mathcal{Y} \in \mathcal{R} ^ {2 \times J}$ for 2D object.

The kinematic function is defined on a kinematic tree. An example is shown in Fig.~\ref{fig:kinematic}. Each joint is associated with a local coordinate transformation defined in the motion parameter, including a rotation from its rotation angles and a translation from its out-coming bones. The final coordinate of a joint is obtained by multiplying a series of transformation matrices along the path from the root joint to itself. Generally, the global position of joint $u$ is
\begin{equation} \label{eq:chain}
p_u = (\prod_{v \in Pa(u)}{Rot(\theta_v) \times Trans(l_v)}) \textbf{O} ^\top
\end{equation}
where $Pa(u)$ is the set of its parents nodes at the kinematic tree, and $\textbf{O}$ is the origin in homogenous coordinate, i.e., $\textbf{O} = [0, 0, 1] ^\top$ for 2D and  $\textbf{O} = [0, 0, 0, 1] ^\top$ for 3D.
For 3D kinematic model, each rotation is assigned with one of the $\{X, Y, Z\}$ axis, and at each joint there can be multiple rotations. The direction of translation is defined in the canonical local coordinate frame where the motion parameters are all zeros.

In ~\cite{zhou2016model}, individual bounds for each angle can be set as additional prior knowledge for the objects. It is feasible for human hand since all the joints have at most 2 rotation angles and their physical meaning is clear. However, in the case of human body, angle constraint are not individual, it is conditioned on pose~\cite{akhter2015pose} and hard to formulate. We leave it as future work to explore more efficient and expressive constraints.

As shown in Fig.~\ref{fig:kinematic}, the forward kinematic function is continuous with respect to the motion parameter. It is thus differentiable. As each parameter occurs in one matrix, this allows easy implementation of back-propagation. We simply replace the corresponding rotational matrix by its derivation matrix and keep other items unchanged. The kinematic model can be easily put in a neural network as a layer for gradient descent-based optimization.

\subsection{Deep Learning with a Kinematic Layer}
\begin{figure}
\begin{center}
\includegraphics[width=0.95\linewidth]{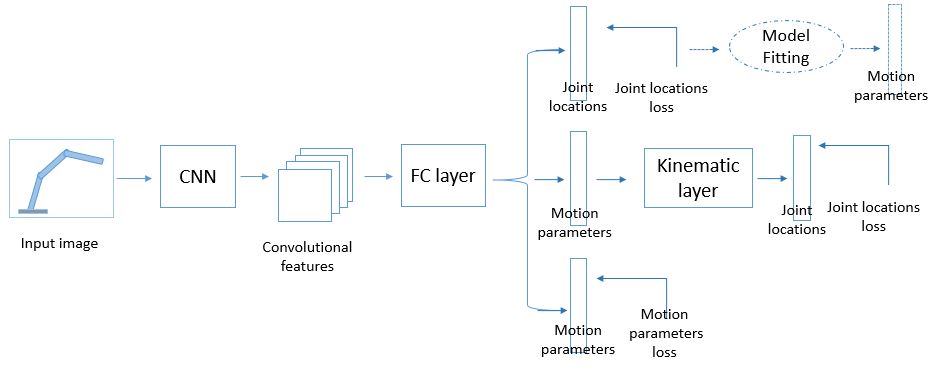}
\end{center}
   \caption{Three methods for object pose estimation. Top (\textbf{Direct Joint}): the network directly outputs all the joints. Such estimated joints could be invalid geometrically. Optionally, they can be optimized via a model-fitting step to recover a correct model, referred to as \textbf{ModelFit} in the text. Middle (\textbf{Kinematic Joint}): our proposed approach. The network outputs motion parameters to the kinematic layer. The layer outputs joints. Bottom (\textbf{Direct Parameter}): the network directly outputs motion parameters.}
\label{fig:methods}
\end{figure}

We discuss our proposed approach and the other two baseline methods to learn the pose of an articulated object. They are illustrated in Fig.~\ref{fig:methods}. All three methods share the same basic convolutional neutral network and only differs in their ending parts, which is parameter-free. Therefore, we can make fair comparison between the three methods.

Now we elaborate on them. The first method is a baseline. It directly estimates the joint locations by a convolutional neural network, using Euclidean Loss on the joints. It is called ~\textbf{direct joint}. It has been used for human pose estimation~\cite{toshev2014deeppose,li20143d} and hand pose estimation ~\cite{oberweger2015hands}. This approach does not consider the geometry constraints of the object. The output is less structured and could be invalid, geometrically.

Instead, we propose to use a kinematic layer at the top of the network. The network predicts the motion parameters of the object, while the learning is still guided by the joint location loss. We call this approach~\textbf{kinematic joint}. The joint location loss with respect to model parameter $\Theta$ is Euclidean Loss
\begin{equation} \label{eq:loss}
L(\Theta) = \frac{1}{2}||\mathcal{F}(\Theta) - Y||^2
\end{equation}
where $Y \in \mathcal{Y}$ is the ground truth joint location in the input image. Since this layer has no free parameters to learn and appears in the end of the network, we can think of the layer as coupled with the Euclidean loss Layer, serving as a geometrically more accurate loss layer.

Compared to direct joint approach, our proposed method fully incorporates prior geometric knowledge of the object, such as the bone lengths and spatial relations between the joints. The joint location is obtained by a generative process and guaranteed to be valid. The motion parameter space is more compact than the unconstrained joint space, that is, the degrees of freedom of motion parameters are smaller than that of joints, for example, in Section 4.2, the DOF is 27 for motion parameters but 51 for joints. Overall, our method can be considered as a better regularization on the output space.

Unlike dictionary-based representations~\cite{Wang_2014_CVPR,Zhou_2015_CVPR} that require a heuristic sparse regularization, our approach has a clear geometrical interpretation and its optimization is feasible in deep neutral network training. Besides, it produces joint rotation angles that could be useful in certain applications.

The third method is a less obvious baseline. It directly estimates the motion parameters, using Euclidean loss on those parameters. It is called~\textbf{direct parameter}. Intuitively, this approach cannot work well because the roles of different parameters are quite different and it is hard to balance the learning weights between those parameters. For example, the global rotation angles on the root joint affects all joints. It has much more impacts than those parameters on distal joints but it is hard to quantify this observation. Moreover, for complex articulated objects the joint locations to joint angles mapping is not one-to-one but ambiguous, e.g., when the entire arm is straight, roll angle on the shoulder joint can be arbitrary and it does not affect the location of elbow and wrist. It is hard to resolve such ambiguity in the network training. By contrast, the joint location loss in our kinematic approach is widely distributed over all object parts. It is well behaved and less ambiguous.

We note that it is possible to enforce the geometric constraints by fitting a kinematic model to some estimated joints as a post-processing~\cite{tompson14tog,Zhou_2016_CVPR}. For example, ~\cite{tompson14tog} recovers a 3D kinematic hand model using a PSO-based optimization, by fitting the model into the 2D hand joint heat maps. ~\cite{Zhou_2016_CVPR} obtains 3D human joints represented by a sparse dictionary using an EM optimization algorithm. In our case, we provide an additional~\textbf{ModelFit} baseline that recovers a kinematic model from the output of direct joint baseline by minimizing the loss in Eq.~\ref{eq:loss}.

\section{Experiment}

\begin{figure}
\begin{center}
\includegraphics[width=0.45\linewidth]{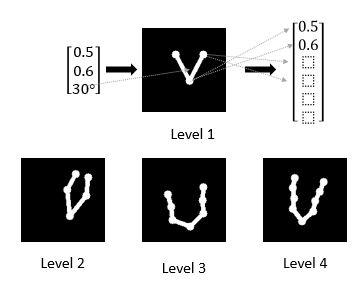}
\end{center}
   \caption{Illustration of the toy problem. The input images are synthesized and binary. \textbf{Top}: Motion parameter and joint representation of a simple object with 3 motion parameters. \textbf{Bottom}: Example input images for 3 objects with different complexity levels. They have 6, 8, and 10 motion parameters, respectively.}
\label{fig:toy}
\end{figure}

The work in ~\cite{zhou2016model} applies the kinematic pose regression approach for depth based 3D hand pose estimation and has shown good results. To verify the generality of the idea, we apply this approach for two more different problems. The first is a toy example for simple 2D articulated object on synthesized binary image. The second is 3D human pose estimation from single RGB images, which is very challenging.

\subsection{A Toy Problem}

In the toy problem, the object is 2D. The image is synthesized and binary. As shown in Fig. ~\ref{fig:toy} top, the input image is generated from a 3 dimensional motion parameter $\Theta = \{x, y, \theta\}$, where $x, y$ is the image coordinate (normalized between $0 - 1$) of the root joint, and $\theta$ indicates the angle between the each bone and the vertical line.

We use a 5 layer convolutional neutral network. The network structure and hyper-parameters are the same as ~\cite{zhou2016model}. The input image resolution is $128 \times 128$. The bone length is fixed as $45$ pixels. We randomly synthesize $16k$ samples for training and $1k$ samples for testing. Each model is trained for $50$ epoches.

\begin{figure}
\begin{center}
\includegraphics[width=0.45\linewidth]{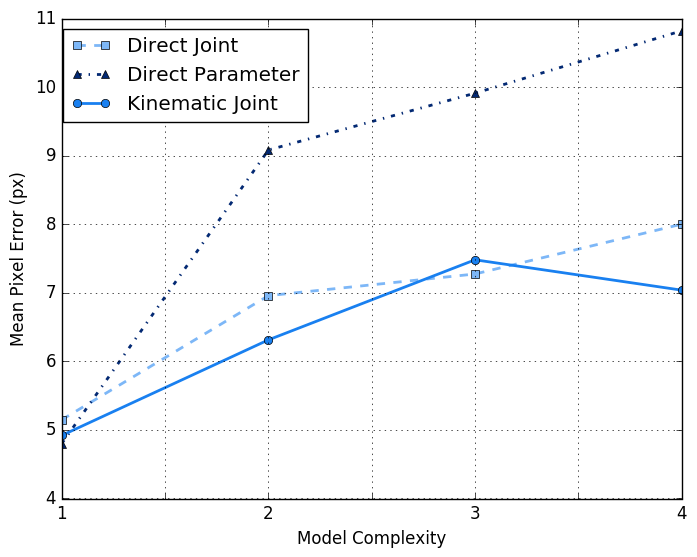}
\includegraphics[width=0.45\linewidth]{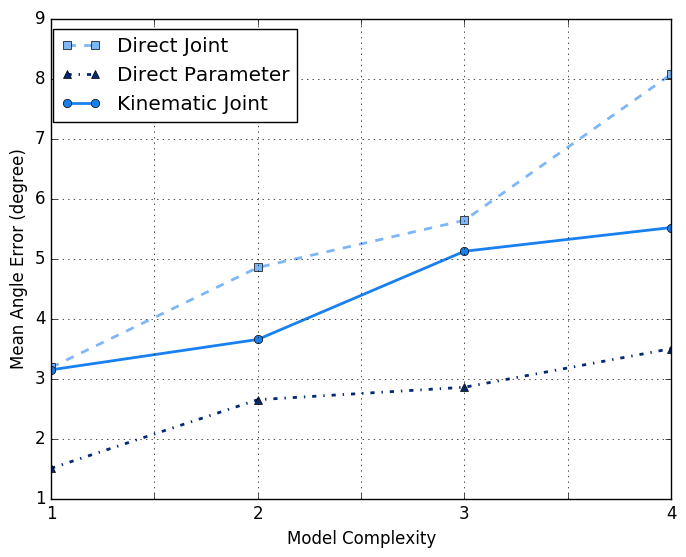}
\end{center}
   \caption{Experimental results on mean joint locations error(\textbf{Left}) and mean angle error(\textbf{Right}) with respect to model complexity. It shows when as kinematic model becoming complex, our approach is stable in both metric.}
\label{fig:toycomp}
\end{figure}

As described in Fig.~\ref{fig:methods}, we perform our \textbf{direct joint}, \textbf{kinematic joint} and \textbf{direct parameter} on this task. The joint location for \textbf{direct parameter} is computed by the kinematic layer as a post process in testing. It turns out all the 3 methods achieve low joint errors in this simple case. The mean joint errors for \textbf{direct joint}, \textbf{kinematic Joint}, \textbf{direct parameter} are 5.1 pixels, 4.9 pixels, and 4.8 pixels, respectively.
\textbf{direct joint} is the worst, probably because the task is easy for all the setting and these two require to learn more parameters. When we evaluate the average length of the two bones for \textbf{direct joint} regression, we find it has a standard deviation of $5.3$ pixels ($11.8\%$ of the bone length 45 pixels), indicating that the geometry constraint is badly violated.

Since it is hard to claim any other significant difference between the 3 method in such a simple case, we gradually increase the model complexity. Global orientation and more joint angles are added to the kinematic model. For each level of complexity, we add one more bone with one rotational angle on each distal bone. Example input image are illustrated in Fig.~\ref{fig:toy} bottom.

The joint location errors and angle errors with respect to the model complexity are shown in Fig.~\ref{fig:toycomp}. Note that for \textbf{direct joint} regression, the angles are directly computed from the triangle. The results show that the task become more difficult for all methods.
\textbf{Direct parameter} gets high joint location errors, probably because a low motion parameter error does not necessarily implies a low joint error. It is intuitive that it always get best performance on joint angle, since it is the desired learning target.
\textbf{Direct joint} regression also has large error on its recovered joint angles, and the average length of each bone becomes more unstable. It shows that geometry structure is not easy to learn. Using a generative \textbf{kinematic joint} layer keeps a decent accuracy on both metric among all model complexity. This is important for complex objects in real applications, such as human body.

\subsection{3D Human Pose Regression}

\begin{figure}
\begin{center}
\includegraphics[width=0.3\linewidth]{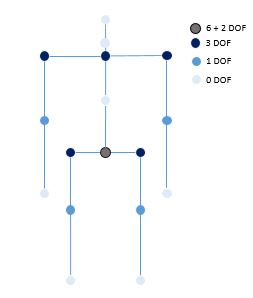}
\end{center}
   \caption{Illustration of Human Model. It contains 17 joints and 27 motion parameters. See text for the detail kinematic structure.}
\label{fig:humanmodel}
\end{figure}

We test our method on the problem of full 3D human pose estimation from single view RGB images. Following ~\cite{li20143d}, the 3D coordinate of joints is represented by its offset to a root joint. We use Human 3.6M dataset~\cite{h36m_pami}. Following the standard protocol in ~\cite{h36m_pami,li20143d,Zhou_2015_CVPR}, we define $J = 17$ joints on the human body. The dataset contains millions of frames of RGB images. They are captured over 7 subjects performing 15 actions from 4 different camera views. Each frame is accurately annotated by a MoCap system. We treat the 4 cameras of the same subject separately. The training and testing data partition follows previous works~\cite{h36m_pami,li20143d,Zhou_2016_CVPR}. All frames from 5 subjects(S1, S5, S6, S7, S8) are used for training. The remaining 2 subjects(S9, S11) are for testing.

Our kinematic human model is illustrated in Fig.~\ref{fig:humanmodel}. It defines $17$ joints with $27$ motion parameters. The pelvis is set as the root joint. Upside it is the neck, which can roll and yaw among the root. Torso is defined as the mid point of neck and pelvis. It has no motion parameter. Pelvis and neck orientation determine the positions of shoulders and hips by a fixed bone transform. Each shoulder/hip has full 3 rotational angles, and elbow/knee has 1 rotational angle. Neck also has 3 rotational angles for nose and head orientation. Note that there can be additional rotation angles on the model, for example shoulders can rotate among neck within a subtle degree and elbows can roll itself. Our rule of thumb is to simulate real human structure and keep the model simple.

We found that the ground truth 3D joints in the dataset has strictly the same length for each bone across all the frames on the same subject. Also, the lengths of the same bone across the 7 subjects are very close. Therefore, in our human model, the bone lengths are simply set as the average bone lengths of the 7 subjects. In addition, every subject is assigned a global scale. The scale is computed from the sum bone lengths divided by the average sum bone length. It is a fixed constant for each subject during training. During testing, we assume the subject scale is unknown and simply set it as 1. In practical scenarios, the subject scale can be estimated by a calibrating pre processing and then fixed.

\setlength{\tabcolsep}{2pt}
\begin{table}

\scriptsize
\begin{center}
\begin{tabular}{ccccccccc}
\hline\noalign{\smallskip}
 & Directions & Discussion & Eating & Greeting & Phoning & Photo & Posing & Purchases\\
\noalign{\smallskip}
\hline
\noalign{\smallskip}
LinKDE~\cite{h36m_pami} & 132.71 & 183.55 & 132.37 & 164.39 & 162.12 & 205.94 & 150.61 & 171.31\\
Li et al~\cite{li20143d} & - & 148.79 & 104.01 & 127.17 & - & 189.08 & - & - \\
Li et al~\cite{Li_2015_ICCV} & - & 136.88 & 96.94 & 124.74 & - & 168.68 & - & - \\
Tekin et al~\cite{tekin2016structured} & - & 129.06 & 91.43 & 121.68 & - & 162.17 & - & - \\
Tekin et al~\cite{Tekin_2016_CVPR} & 132.71 & 158.52 & 87.95 & 126.83 & 118.37 & 185.02 & 114.69 & 107.61\\
Zhou et al~\cite{Zhou_2016_CVPR} & \bf 87.36 & 109.31 & \bf 87.05 & 103.16 & 116.18 & 143.32 & 106.88 & 99.78\\
Ours(Direct) & 106.38 & 104.68 & 104.28 & 107.80 & 115.44 & \bf 114.05 & 103.80 & 109.03 \\
Ours(ModelFit) & 109.75 & 110.47 & 113.98 & 112.17 & 123.66 & 122.82 & 121.27 & 117.98 \\
Ours(Kinematic) & 91.83 & \bf 102.41 & 96.95 & \bf 98.75 & \bf 113.35 & 125.22 & \bf 90.04 & \bf 93.84 \\
\hline\noalign{\smallskip}
 & Sitting & SittingDown & Smoking & Waiting & WalkDog & Walking & WalkPair & Average\\
\noalign{\smallskip}
\hline
\noalign{\smallskip}
LinKDE~\cite{h36m_pami} & 151.57 & 243.03 & 162.14 & 170.69 & 177.13 & 96.60 & 127.88 & 162.14\\
Li et al~\cite{li20143d} & - & - & - & - & 146.59 & 77.60 & - & - \\
Li et al~\cite{Li_2015_ICCV} & - & - & - & - & 132.17 & 69.97 & - & - \\
Tekin et al~\cite{tekin2016structured} & - & - & - & - & 130.53 & \bf 65.75 & - & - \\
Tekin et al~\cite{Tekin_2016_CVPR} & 136.15 & 205.65 & 118.21 & 146.66 & 128.11 & 65.86 & \bf 77.21 & 125.28\\
Zhou et al~\cite{Zhou_2016_CVPR} &\bf 124.52 & 199.23 & 107.42 & 118.09 & 114.23 & 79.39 & 97.70 & 113.01\\
Ours(Direct) & 125.87 & \bf 149.15 & 112.64 & 105.37 & \bf 113.69 & 98.19 & 110.17 & 112.03 \\
Ours(ModelFit) & 137.29 & 157.44 & 136.85 & 110.57 & 128.16 & 102.25 & 114.61 & 121.28 \\
Ours(Kinematic) & 132.16 & 158.97 & \bf 106.91 & \bf 94.41 & 126.04 & 79.02 & 98.96 & \bf 107.26 \\
\hline
\end{tabular}
\caption{Results of Human3.6M Dataset. The numbers are mean Euclidean distance(mm) between the ground-truth 3D joints and the estimations of different methods.}
\label{table:H36M}
\end{center}
\end{table}
\setlength{\tabcolsep}{1.4pt}

Following ~\cite{li20143d,tekin2016structured}, we assume the bounding box for the subject in known. The input images are resized to $224 \times 224$. Note that it is important not to change the aspect ratio for the kinematic based method, we use border padding to keep the real aspect ratio. The training target is also normalized by the bounding box size. Since our method is not action-dependent, we train our model using all the data from the 15 actions. By contrast, previous methods ~\cite{h36m_pami,Li_2015_ICCV,Zhou_2016_CVPR} use data for each action individually, as their local feature, retrieval database or pose dictionary may prefer more concrete templates.

We use the 50-layer Residual Network~\cite{He_2016_CVPR} that is pre-trained on ImageNet~\cite{DBLP:journals/corr/RussakovskyDSKSMHKKBBF14} as our initial model. It is then fine-tuned on our task. Totally available training data for the $5$ subjects is about 1.5 million images. They are highly similar and redundant. We randomly sample 800k frames for training. No data augmentation is used. We train our network for 70 epoches, with base learning rate 0.003 (dropped to 0.0003 after 50 epochs), batch size 52 (on 2 GPUs), weight decay 0.0002 and momentum 0.9. Batch-normalization~\cite{ioffe2015batch} is used. Our implementation is based on Caffe~\cite{jia2014caffe}.

The experimental results are shown in Table~\ref{table:H36M}. The results for comparison methods~\cite{h36m_pami,li20143d,Li_2015_ICCV,tekin2016structured,tekin2016structured,Tekin_2016_CVPR,Zhou_2016_CVPR} are from their published papers. Thanks to the powerful Residual Network~\cite{He_2016_CVPR}, our \textbf{direct joint} regression base line is already the state-of-the-art. Since we used additional training data from ImageNet, comparing our results to previous works is unfair, and the superior performance of our approach is not the contribution of this work. We include the previous works' results in Table~\ref{table:H36M} just as references.

\textbf{Kinematic joint} achieves the best average accuracy among all methods, demonstrating that embedding a kinematic layer in the network is effective. Qualitative results are shown in Table~\ref{table:demo}, including some typical failure cases for \textbf{direct joint} include flipping the left and right leg when the person is back to the camera(Row 1) and abnormal bone length(Row 2,3).

\begin{figure}
\begin{center}
\includegraphics[width=0.55\linewidth]{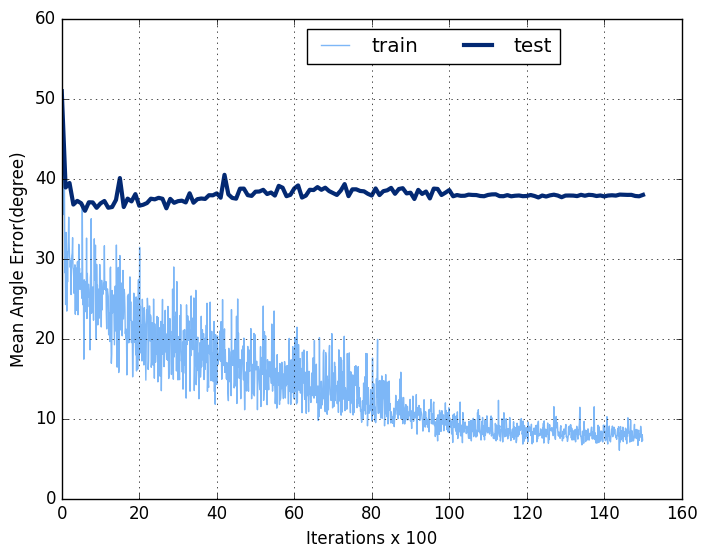}
\end{center}
   \caption{Training curve of direct motion parameter regression. Although the training loss keeps dropping, the testing loss remains high. }
\label{fig:paramcurve}
\end{figure}

Despite \textbf{direct joint} regression achieve a decent accuracy for 3D joint location, we can further apply a kinematic model fitting step, as described in the previous sections. The model fitting is based on gradient-descent for each frame. The results is shown in Table.~\ref{table:H36M} as \textbf{ours(Fit)}, it turns out to be worse than \textbf{direct joint}, indicating such post-preprocessing is sub-optimal if the initial poses do not have valid structural information.

We also tried ~\textbf{direct parameter} regression on this dataset. The training target for motion parameter is obtained in the same way as described above, by gradient descent. However, as shown in Fig.~\ref{fig:paramcurve}, the testing error keeps high. Indicating direct parameter regression does not work on this task. There could be two reasons: many joints have full 3 rotational angles, this can easily cause ambiguous angle target, for example, if the elbow or knee is straight, the roll angle for shoulder or hip can be arbitrary. Secondly, learning 3D rotational angles is more obscure than learning 3D joint offsets. It is even hard for human to annotate the 3D rotational angles from an RGB image. Thus it may require more data or more time to train.

\begin{table}
     \begin{center}
     \begin{tabular}{ c c c c }

      Input Image & Direct Joint & Kinematic Joint& Ground-truth\\ 
      \\
     \includegraphics[width=0.22\textwidth]{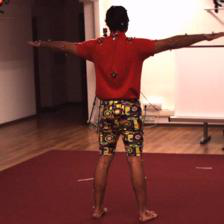}
      &\includegraphics[width=0.25\textwidth]{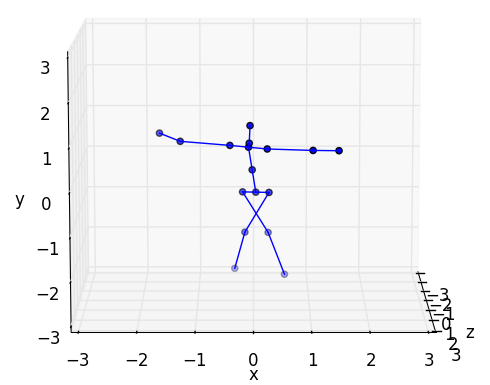}
      &\includegraphics[width=0.25\textwidth]{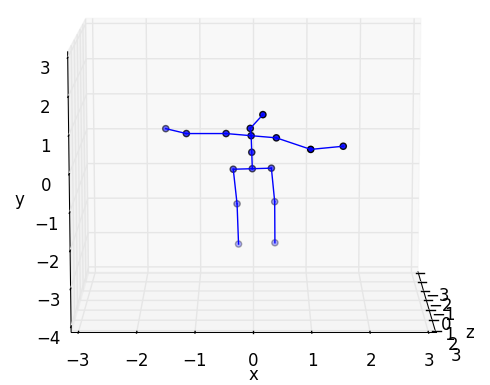}
      &\includegraphics[width=0.25\textwidth]{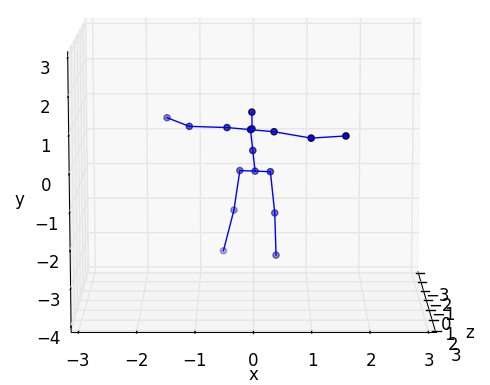}
      \\ 
     \includegraphics[width=0.22\textwidth]{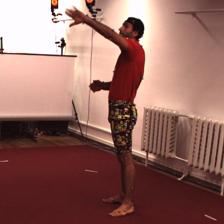}
      &\includegraphics[width=0.25\textwidth]{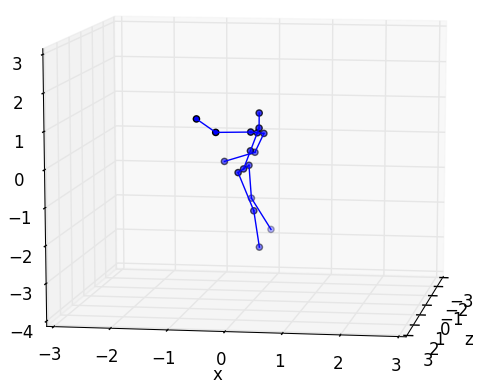}
      &\includegraphics[width=0.25\textwidth]{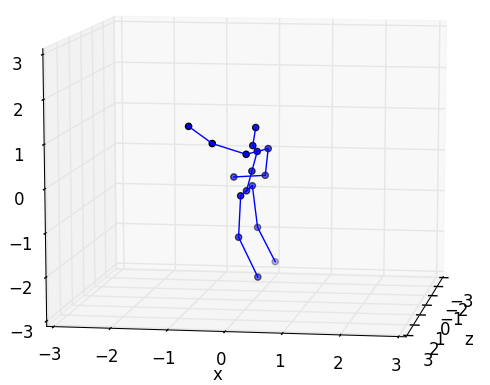}
      &\includegraphics[width=0.25\textwidth]{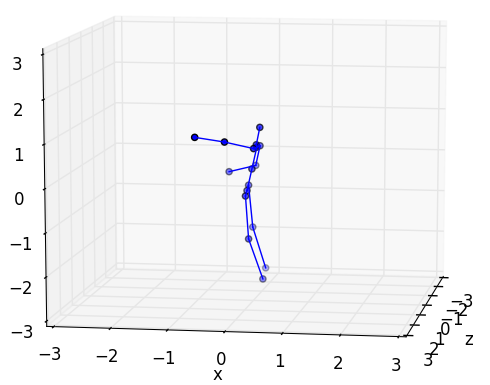}
      \\ 
     \includegraphics[width=0.22\textwidth]{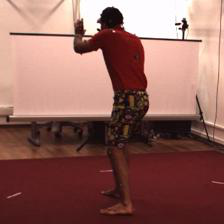}
      &\includegraphics[width=0.25\textwidth]{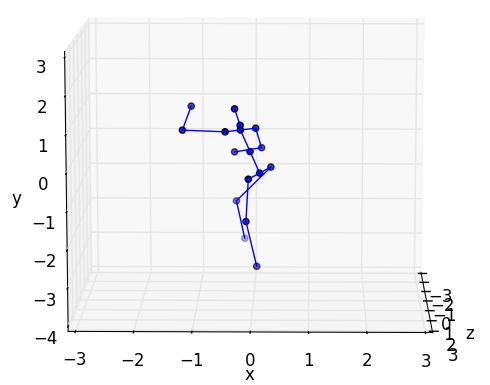}
      &\includegraphics[width=0.25\textwidth]{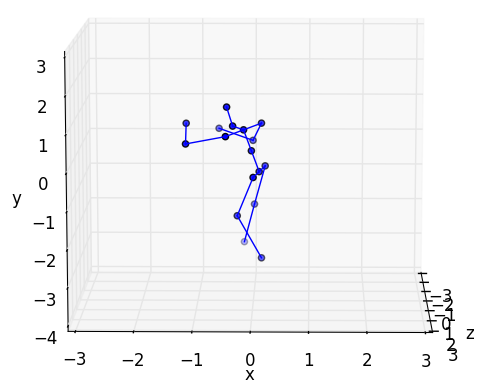}
      &\includegraphics[width=0.25\textwidth]{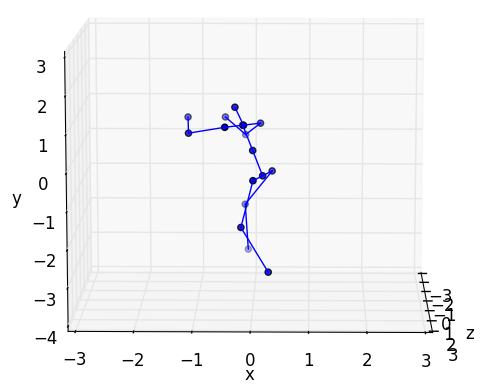}
      \\ 
     \includegraphics[width=0.22\textwidth]{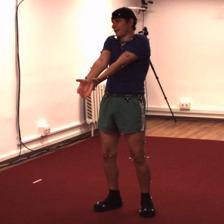}
      &\includegraphics[width=0.25\textwidth]{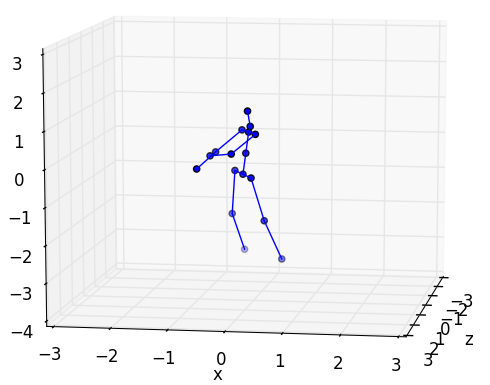}
      &\includegraphics[width=0.25\textwidth]{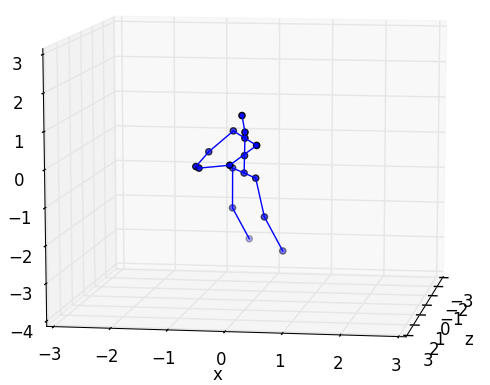}
      &\includegraphics[width=0.25\textwidth]{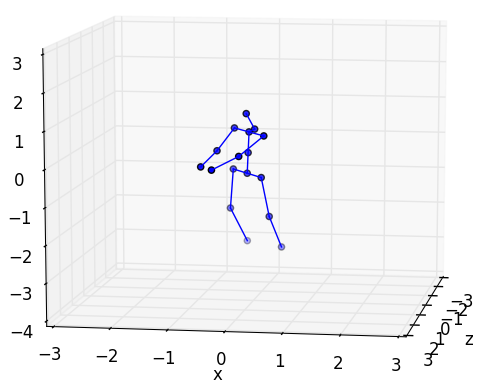}
      \\ 
     \includegraphics[width=0.22\textwidth]{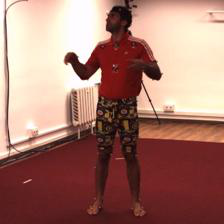}
      &\includegraphics[width=0.25\textwidth]{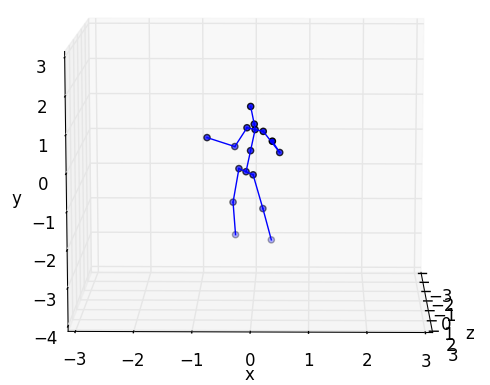}
      &\includegraphics[width=0.25\textwidth]{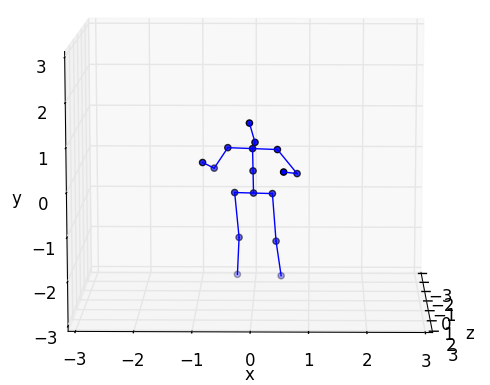}
      &\includegraphics[width=0.25\textwidth]{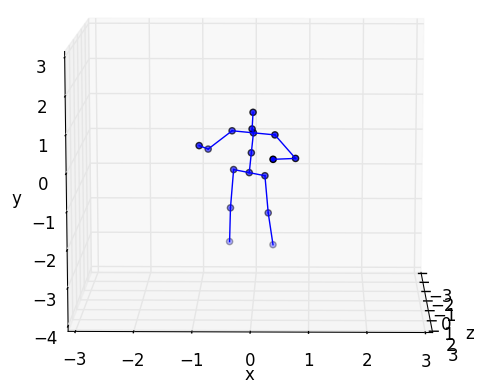}
      \\ 
     \includegraphics[width=0.22\textwidth]{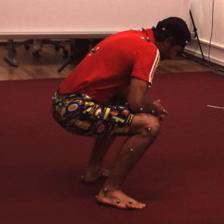}
      &\includegraphics[width=0.25\textwidth]{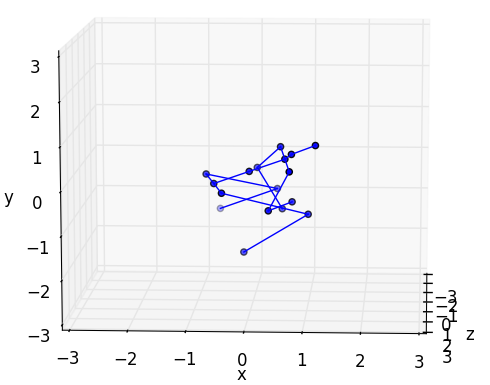}
      &\includegraphics[width=0.25\textwidth]{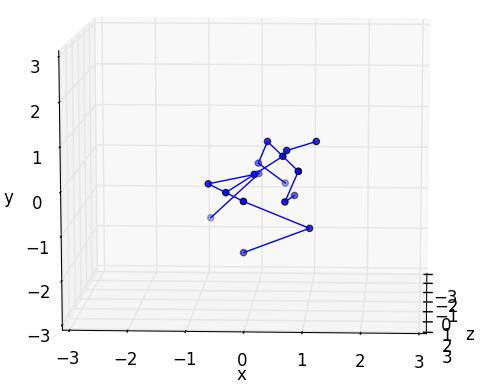}
      &\includegraphics[width=0.25\textwidth]{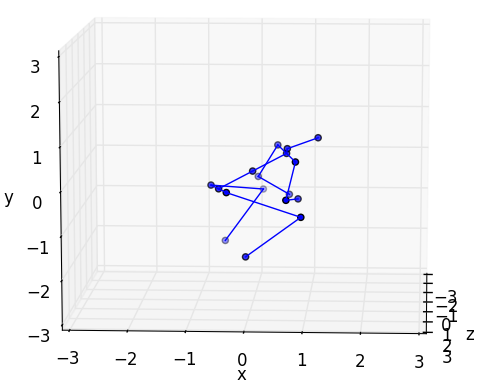}
      \\ 
      \end{tabular}
      \caption{Qualitative results for direct joint regression and kinematic on Human3.6M dataset. They show some typical characters for these methods.The results are ploted at 3D space from the same viewpoint.}
      \label{table:demo}
      \end{center}
\end{table}


\section{Conclusions}
We show that geometric model of articulated objects can be effectively used within the convolutional neural network. The learning is end-to-end and we get rid of the inconvenient post-processing as in previous approaches. The experimental results on 3D human pose estimation shows that our approach is effective for complex problems. In the future work, we plan to investigate more sophisticated constraints such as those on motion parameters. We hope this work can inspire more works on combining geometry with deep learning.

\section*{Acknowledgments}
We would like to thank anonymous reviewers who gave
us useful comments. This work was supported by Natural
Science Foundation of China (No.61473091), National Science Foundation of China (No.61305091), and The Fundamental Research Funds for the Central Universities (No.2100219054).

\bibliographystyle{splncs03}
\bibliography{egbib}
\end{document}